# Expectation-based Minimalist Grammars

**Cristiano Chesi**
NeTS – IUSS lab for NEurolinguistics, Computational Linguistics,
and Theoretical Syntax, Pavia
cristiano.chesi@iusspavia.it

## Abstract

Expectation-based Minimalist Grammars (e-MGs) are simplified versions of the (Conflated) Minimalist Grammars, (C)MGs, formalized by Stabler (Stabler, 2011, 2013, 1997) and Phase-based Minimalist Grammars, PMGs (Chesi, 2005, 2007; Stabler, 2011). The crucial simplification consists of driving structure building only using lexically encoded categorial top-down expectations. The commitment on a top-down procedure (in e-MGs and PMGs, as opposed to (C)MGs, Chomsky, 1995; Stabler, 2011) allows us to define a core derivation that is the same in both parsing and generation (Momma & Phillips, 2018).

## 1 Introduction

Minimalism (Chomsky, 1995, 2001) is an elegant transformational grammatical framework that defines structural dependencies in phrasal (i.e. hierarchical) terms simply relying on one core structure building operation, Merge, that combines lexical items and the result of other Merge operations. (1).a is the representative result of two ordered Merge operations (i.e. Merge(γ, Merge(α, β)) both taking the items α, β and γ directly from the lexicon, while (1).b relies on the so called Internal Merge (Move): the re-Merge of an item that was already merged in the structure.

(1) a. [γ [α, β]] *Merge* only
b. [β [γ [α, $__{\beta}$]] *Merge + Move*

As result, Move connects the item at the edge of the structure (β) with a trace ($__{\beta}$), a phonetically empty copy of the item that in a previous Merge operation combined with a hierarchically lower item (α in (1).b). In both (Conflated) Minimalist and Phase-based Minimalist Grammars ([C]MGs and PMGs respectively) Merge and Move are feature-driven operations, that is, a successful operation must be triggered by the relevant (categorial) features matching, and, once these features are used, they get deleted. Consequently, a feature pair is always responsible for each operation (unless specific features are left unerased after a successful operation, as in raising predicates and successive cyclic movement, Stabler 2011). One crucial difference between PMGs and MGs is that while MGs operate from-bottom-to-top, as indicated in (2), PMGs structure building operations apply top-down as schematized in (3)[1]:

(2) $Merge(\alpha_{=X}, {}_{X}\beta) = [\alpha\ [\alpha_{\cancel{=X}}\ {}_{\cancel{X}}\beta]]$ *MGs*
$Move({}_{+Y}\alpha, [\ldots\ \beta_{-Y}\ \ldots]) =$
$[\alpha\ [\beta_{\cancel{-Y}}\ [{}_{\cancel{+Y}}\alpha\ [\ldots\ \beta_{-Y}\ \ldots]]]]$

(3) $Merge(\alpha_{=X}, {}_{X}\beta) = [\alpha_{\cancel{=X}}\ [{}_{\cancel{X}}\beta]]$ *PMGs*
$Move([\alpha_{=S\ \cancel{+Y}}[{}_{\cancel{Y}\ Z}\ \beta]]) =$
$[\alpha_{\cancel{=S}\ \cancel{+Y}}[{}_{\cancel{Y}\ Z}\ \beta]\ {}_{S}[\ldots\ ({}_{=Z}\ [{}_{Z}\ \beta])\ \ldots]]$

Another relevant difference between the two approaches is related to the implementation of Move: MGs use the "+/-" feature distinction and the same deletion procedure after matching, while PMGs do not use "-" features and simply assume that both "+" and "=" select categorial features, which are deleted after Merge. In PMGs, "+" features force memory storage and hence the movement (downward) of the licensed item, until the relevant prominent category identifying the moved item (*Z* in (3)) is selected. If no proper selection is found, the sentence is ungrammatical. CMG as well dispenses the grammar with the +/- feature distinction and only relies on select features (=*X*), but it must assume that feature deletion can be procrastinated (again, for instance, in raising predicates). Despite the fact that, from a generative point of view, all these formalisms are equivalent and they all fall under the so called mildly-context sensitive domain (Stabler, 2011), it is worth to appreciate the dynamics of structure building "on-line", namely how the derivation unrolls, word by word. Taking the MGs lexicon (4), the expected constituents in (1) are built adding items to the left-edge of the structure at each Merge/Move application, as described in (5).

[1] *α* and *β* are lexical items, =*X* indicates the selection of *X*, where X is a categorial feature. Lexical items are tuples consisting of selections/expectations (=*X*) and categories (X, i.e. selected/expected features); for convenience, select features are expressed by rightward subscripts, and categories as leftward subscripts. Similarly, Move is driven by licensing (-*Y*, leftward subscripts) and licensors (+*Y*, rightward subscripts) features (Stabler, 2011).

(4) $Lex_{MG} = \{[_{Y}\alpha_{=X}], [_{X\ -Z}\beta], [\gamma_{=Y\ +Z}]\}$

(5) i. *Merge*($_{Y}\alpha_{=X}$, $_{X\ -Z}\beta$) = $[_{Y}\alpha\ [\alpha_{\cancel{=X}}\ _{\cancel{X}\ -Z}\beta]]$

ii. *Merge*($\gamma_{=Y\ +Z}$, $[_{Y}\alpha\ [\alpha\ _{-Z}\beta]]$) =
$[\gamma_{\cancel{=Y}\ +Z}\ [_{\cancel{Y}}\alpha\ [\alpha\ \ _{-Z}\beta]]]$

iii. *Move* ($[\gamma_{+Z}\ [\alpha\ [\alpha\ _{-Z}\beta]]]$) =
$[[_{\cancel{-Z}}\beta]\ \gamma_{\cancel{+Z}}\ [\gamma\ [\alpha\ [\alpha\ __{\beta}]]]]$

An equivalent structure is obtained in PMGs[2] as shown in (7). Notice a minimal difference in the lexicon (6): the absence of the "-" features.

(6) $Lex_{PMG} = \{\ [_{Y}\alpha_{=X}], [_{Z\ X}\ \beta], [\gamma_{+Z\ =Y}]\ \}$

(7) i. *Merge*($_{Z\ X}\beta$, $\gamma_{+Z\ =Y}$) = $[[_{\cancel{Z}\ X}\beta]\ \gamma_{\cancel{+Z}\ =Y}]$
$_{X}\beta \longrightarrow M$

ii. *Merge*($[[\beta]\ \gamma\ _{=Y}]$, $_{Y}\alpha\ _{=X}$) =
$[[\beta]\ \gamma\ [_{(\gamma)}\ _{\cancel{=Y}}\ [_{\cancel{Y}}\alpha\ _{=X}]]]$ $M = \{_{X}\beta\}$

iii. *Move*($[[\beta]\ \gamma\ [_{(\gamma)}\ [\alpha\ _{=X}]]$, $_{X}\beta$) =
$[[\beta]\ \gamma\ [_{(\gamma)}\ [\alpha_{\cancel{=X}}\ [_{(\alpha)}\ _{\cancel{X}}__{\beta}]]]]$ $_{X}\beta \longleftarrow M$

The result of the two derivations is (strongly) equivalent in hierarchical (and dependency) terms. The simplicity, in pre-theoretical terms, of the two descriptions is comparable: while PMGs must postulate the *M* storage to implement Move (as result of the missing selection of a categorial feature), MGs must postulate an independent workspace to build nontrivial left-branching structures, for instance before merging a multi-word subject like "the boy" with its predicate (e.g., "runs"). Furthermore, both formalisms must restrict the behavior either of the *M* buffer operativity or the accessibility to the *-f* features to limits the Move operation (e.g., island constraints, Huang, 1982).

### 1.1 Top-down is Better

There are at least three reasons to commit ourselves to the top-down orientation instead of remaining agnostic or relying on the mainstream Minimalist brick-over-brick (from-bottom-to-top) approach (Chesi, 2007): First, the order in which the structure is built is grossly transparent with respect to the order in which the words are processed in real-life tasks, both in generation and in parsing in PMGs, but not in MGs.

Second, in PMGs, the simple processing order of multiple expectations is sufficient to distinguish between sequential (the last expectation of a given lexical item) and nested expectations (any other expectation): The first qualifies as the transparent branch of the tree (i.e. it is able to license pending items from the superordinate selecting item), while constituents licensed by nested expectations qualify as configurational islands (Bianchi & Chesi, 2006; Chesi, 2015). Moreover, successive cyclic movement is easily described in PMGs without relying on feature checking at any step or non-deterministic assumptions on features deletion (Chesi, 2015) contrary to (C)MGs.

A third logical reason to prefer the top-down orientation over the bottom-up alternative is related to the unicity of the root node in tree graphs and it deserves a specific section (§1.2).

### 1.2 Single Root Condition

Another logical reason to prefer the top-down orientation over the bottom-up alternative is related to the unicity of the root node. As anticipated, the creation of complex (binary) branching structures poses a puzzle for (C)MGs: independent workspaces must be postulated, namely [the boy] and the [sings … ] phrases must be created before the first can merge with the second:

(8) $[_{VP}\ [_{DP}$ the boy$]\ [_{V}$ sings $[_{DP}$ a song$]]$

This is the case for any "complex" subject or adjunct (i.e., non-projecting constituents which are simply composed by multiple words) that must be the result of (at least) one independent Merge operation, before this can merge with the relevant predicate (e.g. $[_{V}$ sings …$]$[3] in (8)). The processing of these constituents represents a major difference between MGs and PMGs derivations.

While MGs must decide where to start from (and both solutions are possible and forcefully logically independent from parsing or generation, which undeniably proceed incrementally "from left to right"), PMGs take advantage of the "single root condition" (Partee et al., 1993, p. 439) and avoid this problem:

[2] Move is implemented using a Last-In-First-Out addressable memory buffer M, where the item (*β*) with unselected categorie(s) (*X*) is stored ("$_{X}\beta \rightarrow M$") and retrieved ("$_{X}\beta \leftarrow M$") when selected (i.e. "=*X*").

[3] Considering the inflection "-s" as part of the lexical element or by (head) moving the root "sing-" to T is uninfluential here. This sort of head movement will be lexically implemented in e-MGs (e.g. $[_{T\ V}$eats $_{=V}$ …$]$.

(9) In every well-formed constituent structure tree, there is exactly one node that dominates every node.

As indicated in (3), the binary operation Merge, simply produce an hierarchical dependency in which the dominating (asymmetrically C-commanding, in the sense of Kayne 1994) item, is above the dominated (C-commanded) one. This is compatible with Stabler notation (10).a-b and plainly solves the ambiguity of the nature of the "label" of the constituent (Rizzi, 2016). In this sense, PMGs (and e-MGs) can adopt directly a more concise description, that is (10).c, totally transparent with respect to the (Universal) Dependency approach (Nivre et al., 2017).

(10) a. MGs b. (C)MGs c. (P/e-)MGs

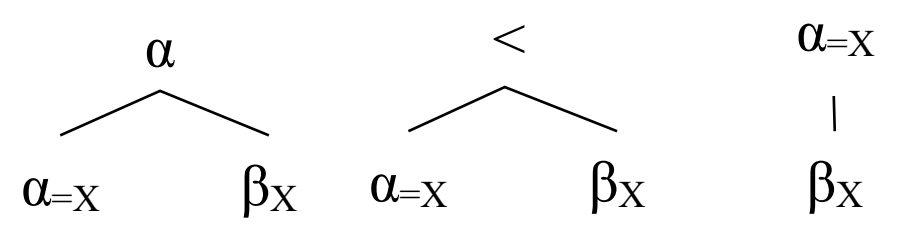

The higher node (possibly the root) is always a selecting item (a *probe*, in minimalist terms), and it is the first item to be processed. This does not necessarily imply that this item is linearized before the selected category (the *goal*, in minimalist terms): if the selecting node has multiple selections, it must remain to the right-edge of the structure to license, locally, the other(s) selection expectation(s). E.g., if $[\alpha_{=X\ =Y}]$, $[_X\beta]$ and $[_Y\gamma]$, then:

(11) $[\alpha_{\cancel{=X}\ =Y}\ [_{\cancel{X}}\beta]\ [(\alpha_{\cancel{=Y}})\ [_{\cancel{Y}}\gamma]]]$

In this case, <α, β, γ> would be the default linearization, but it is easy to derive <β, α, γ> instead, assuming a simple parameterization on spell-out in case of multiple select features (§0).

## 2 The grammar

As (C/P)MGs, e-MGs include a specification of a lexicon (*Lex*) and a set of functions (*F*), the structure building operations. The lexicon, in turn, is a finite set composed by words each consisting of phonetic/orthographic information (*Phon*) and a combination of categorical features (*Cat*), expressing *expect(ations)*, *expected* and *agreement* categories[4]. In the end, an optional set of Parameters (*P*) (see Chesi 2021), inducing minimal modifications to the structure building operations *F* and, possibly, to the *Cat* set, under the fair assumption that *F* and *Cat* are universal. More precisely, any e-MG is a 5-tuple such that:

(12) *G* = (*Phon*, *Cat*, *Lex*, *F*, *P*), where

*Phon*, a finite set of phonetic/orthographic features (i.e., orthographic forms representing words, e.g., "the", "smiles")
*Cat*, a finite set (morphosyntactic categories, that can be *expect*, *expected* or *agreement* features e.g., "D", "V"… "gen(der)", "num(ber)", "pl(ural)" etc.)
*Lex,* a set of expressions built from *Phon* and *Cat* (the lexicon)
*F*, a set of partial functions from tuples of expressions to expressions (the structure building operations)
*P*, a finite set of minimal transformations of *F* and/or *Cat* (the parameters), producing *F'* and *Cat'*, respectively.

### 2.1 Lexical items and categories

Each lexical item *l* in *Lex*, namely each word, is a 4-tuple defined as follows[5]:

(13) l = (*Ph*, *Exp*(*ect*), *Exp*(*ect*)*ed*, *Agr*(*ee*)),

*Phon,* from *Phon* in *G* (e.g., "the")
*Exp,* a finite list of ordered features from *Cat* in *G* (the category/ies that the item expects will follow, e.g., =*N*)
*Exped* is a finite list of ordered features from *Cat* in *G* (the category/ies that should be licensed/expected, e.g., *N*)
*Agr(ee)* is a structured list of features from *Cat* in G (e.g., *gen.fem*, *num.pl*)

All *Exp(ect)*, *Exp(ect)ed* and *Agr(ee)* features are then subsets of *Cat* in G. In *Agr*, for instance, a feminine gender specification (*gen.fem*) expresses a subset relation (i.e., "feminine" ⊆ "gender").

For sake of simplicity, each *l* will be represented as $[_{\text{Expected(; Agree)}}\ \text{Phon}\ _{=/+\text{Expect}}]$ as in (14):

(14) $[_D\,\text{the}\,_{=N}]$, $[_{N;\,num.pl}\,\text{dogs}]$, $[_T\,\text{barks}\,_{=D}]$

[4] As in MGs, lexical items could be specified both for phonetic (*Phon*) and semantic features (*Sem*). In e-MGs, *expectations* (=/+X) and *expectees* (X) correspond to MGs *selectors/licensors* and *selectees/licensees* respectively. *Agreement* features indicate categorial values to be unified (Chesi, 2021).

[5] This is the simplest possible implementation. Attribute-Value Matrices, as in HPSH (Pollard & Sag, 1994) or TRIE/compact trees exploiting the sequence of expectations (Chesi, 2018; Stabler, 2013) are possible implementations.

We refer to the most prominent (i.e., the first) *Expected* feature as the Label (*L*) of the item. E.g., the label *L* of “the” will be *D*, while the label of “barks” will be *T*. Similarly, let us call *S* (for select) the first *Expect* feature and *R* the remaining *Expect(actions)* (if any).

### 2.2 Structure Building operations

Given $l_x$ an arbitrary item such that $l_x = (P_x, L_x/Exped_x, S_x/R_x/Exp_x, Agr_x)$ we can define MERGE as follows:

(15) $\text{MERGE}(l_{1(S1)}, l_{2(L2)}) =$

$$\begin{cases} 1, [l_{1(\cancel{S1})}[l_{2(\cancel{L2})}]] & if\ S_1 = L_2 \\ 0 & otherwise \end{cases}$$

MERGE is implemented as the usual binary function that is successful (it returns “1”) and creates the dependency (asymmetric C-command or inclusion, in set theoretic terms) (10).c, namely [$l_1$ [$l_2$]], if and only if the label of the subsequent item ($l_2$) is exactly the one expected by the preceding item ($l_1$), namely $S_1 = L_2$. This is probably both too strict in one sense (adjuncts are not properly selected) and too permissive in another (certain elements must agree to be merged). In the first case, I assume that [$l_1$ [$l_2$]] can be formed even if $S_1$ is not *=X* but *+X*: while *=X* corresponds to functional selection (in compositional semantics terms Heim & Kratzer 1998), *+X* corresponds to an intersective compositional interpretation (e.g. adjuncts and restrictive relative clauses). As for the agreement constraint, I postulate an extra (possibly parametrized) condition on MERGE, namely the sharing (inclusion) of the relevant *Agr* features associated to some specific categories.

The auxiliary functions necessary to implement Agreement are AGREE and UNIFY and can be minimally defined as follows:

(16) $\text{AGREE}(l_{1(L1)}, l_{2(L2)}) =$

$$\begin{cases} 1 & if\ L_1 \wedge L_2 \in P\{Agr\} \rightarrow Unify(l_1, l_2) \\ 0 & otherwise \end{cases}$$

(17) $\text{UNIFY}(l_{1(agr1)}, l_{2(agr2)}) =$

$$\begin{cases} 1, a, \forall a{:}\, Agr_1 \forall b{:}\, Agr_2\ a \cap b & if\ a \subseteq b \\ 1, b, \forall a{:}\, Agr_1\ \forall b{:}\, Agr_2\ a \cap b & if\ b \subseteq a \\ 0 & otherwise \end{cases}$$

Unification is simply expressed as an inclusion relation returning true and the most specific feature for any possible featural intersection between $l_1$ and $l_2$ *Agr* features[6]. Notice that Agreement is a conditional, parametrized option, that is, it only involves specific categories (possibly specified in the parameter set *P*): if the *L* category belongs to the Agreement set (*Agr*) in *P* for the grammar *G*, unification will be attempted, otherwise agreement will be trivially successful. The fact that AGREE should apply in conjunction with MERGE is straightforward in the D-N domain: in most Romance languages, in which gender and number are shared between the determiner and the noun, we assume that *D* selects *N* (this happens also for intermediate functional specifications, according to the cartographic intuition, Cinque 2002). This is less evident in the Subject – Predicate case, in SV language, where the predicate should select (then precede) *D*. Since the subject is clearly processed (i.e. merged) before *T*, in canonical SV sentences, and it does not select *T*, a re-merge operation should be considered (e.g. case checking). This re-merge (inducing the locality of Agree, *pace* Chomsky 2001) is logically and empirically sound (movement and agreement can be related and parametrized, Alexiadou & Anagnostopoulou 1998). In this case, re-merge must be preceded by MOVE, an operation that stores in memory an item which is “not fully” expected (i.e. there are $exped_2$ features remaining) by the previous MERGE:

(18) $\text{MOVE}(l_{1(M1)}, l_{2(L2)}) =$

$$\begin{cases} 1, Push(M_1, l_{2(Phon_2=\emptyset)}) & if\ L_2 \neq \emptyset \\ 0 & otherwise \end{cases}$$

The definition of MOVE tells us that an item ($l_2$) must be moved (pushed[7]) into the memory buffer ($M_1$) of the superordinate item ($l_1$) if it still has expected features to be selected ($L_2 \neq \emptyset$). Notice that item moved in $M_1$ is not an exact copy of $l_2$: the used features (including *Phon*) will not be stored in memory. This definition produces the expected derivation if it applies right after MERGE, that is, once the item $l_2$ is properly (at least partially) selected; in this case, if $l_2$ still has *exp(ect)ed* features to be licensed, it must hold in the memory buffer of the selecting item, waiting for a proper selection of what has become the new $l_2$ label (i.e. $L_2$). (Re-)Merge is then when agreement will be attempted (i.e. *if* MERGE($l_1$, $l_2$) in §3, should then be interpreted as *if* MERGE($l_1$, $l_2$) $\wedge$ AGREE($l_1$, $l_2$) *then…* for specific parameterized categories). In the end, the top-down derivation in SV languages would unroll as follows: the subject (a DP) is first selected by a superordinate item (presuppositional subject position, situation topic, focus etc.)[8] then it gets (partially) stored in the *M* buffer of the selecting item in virtue of the unselected *D* features, then re-merged as soon as a proper predicate, expressing the relevant *T* category requiring agreement (*T* should be included in the parameterized *Agreement*), is merged and properly selects a *D* argument (or it selects a *V* that later selects *D*). The content of the memory buffer is transmitted (inherited) through the last selected expectation, namely when the expecting and the expectee items successfully merge and the expecting item has no more expectations ($R_1 \neq \emptyset$).

[6] UNIFY(*num*, *num.pl*) = *num.pl*; UNIFY(∅, *num.pl*) = *num.pl*; UNIFY(*gen.f*, *num.pl*) = *gen.f*, *num.pl*, since *gen* and *num* are distinct agree subsets. On the other hand, UNIFY([*gen.f*, *num.sg*], *num.pl*) would fail.

[7] PUSH and POP are trivial functions operating on arrays: insert (PUSH) / remove (POP) an item to/from the first available slot of a stack or a priority queue.

If the expecting item has expectations, then the expected item constitutes a nested expansion, and the inheritance mechanism is blocked:

(19) INHERIT($l_{1(M1)}$, $l_{2(M2)}$) =

$$\begin{cases} 1, M_2 \Leftarrow M_1 & if\ \text{MERGE}(l_1, l_2)\ \wedge R_1 \neq \emptyset) \\ 0 & otherwise \end{cases}$$

The M buffer of the last selected item that does not have other expectations (namely a right phrasal edge, i.e., $S=\emptyset$) must be empty (i.e., $M=\emptyset$). If not, the derivation fails (i.e., it stops) since a pending item remains unlicensed:

(20) SUCCESS($l_{x(Sx, Mx)}$) =

$$\begin{cases} 1, & if\ S_x = \emptyset\ \rightarrow M_x = \emptyset) \\ STOP & otherwise \end{cases}$$

Notice that the sequential item must be properly selected ($=S_X$). If this is not the case, the inheritance would transmit the content of the memory buffer of the superordinate phase into the memory buffer of an adjunct or a restrictive relative clause, which clearly qualify as (right-branching) islands. Therefore, the "restrictive" (since feature driven) MERGE definition in (15) seems correct and empirically more accurate than "free Merge" (Chomsky et al., 2019, p. 238).

# 3 The Derivation Algorithm

We can now define the full-fledged top-down derivation algorithm which is common both to generation and to parsing tasks (§3.2). Consider *cn* to be the *current node*, *exp* the list of pending expectations and *mem* the ordered list of items in memory. We initialize our procedure by picking up an arbitrary node from *G.Lex* as *cn*. Being *cn* the *root* node of our derivation(al tree) and *w* the array of words we want to produce/recognize, we can define the function DERIVE(*cn*, *w*) as follows:

```
while cn.exp & w
  while cn.mem
    foreach cn.mem[i] in cn.mem
      if MERGE(cn.exp[0], cn.mem[i])
          POP(cn.exp)
          POP(cn.mem)
      else break
  if MERGE(cn.exp[0], w[0])
    POP(cn.exp)
    if w[0].exped
      MOVE(cn, w[0])
    if w[0].exp
      cn = w[0]
      INHERIT(exp[0], w[0])
      SUCCESS(w[0])
    POP(w)
    if not cn.exp
      while !cn.exp & (cn != root)
          cn = cn.father
  else fail
```

Informally speaking, as long as we have lexical items to consume (*w*), we loop into the set of expectations of *cn (cn.exp)*, first attempting to Merge items from (*cn.*)*mem* (if any), as in the active filler strategy (Frazier & Clifton, 1989), then consuming words in the input (being *w[0]* the first available word). Remember that each word has *exp(ect)ed* features (the first being the label *L*), *exp(ectations)* and *agr(eement)* features. *Cn*s have their own *mem* that can be inherited only by the last expected item, and, apart from the root node, a *father*. The derivation is then a depth-first, left-right (i.e., real-time) strategy to derive a structure given a grammar, a root node, and a sequence of lexical items to be integrated.

## 3.1 Lexical ambiguity only

Ignoring Parameters, the derivation procedure in §3 should face lexical ambiguity: the same *Phon* in *w[n]* might be associated to multiple items *l* in *Lex* with different features; the default option is to initialize a new derivational tree for any ambiguous item in *Lex*. Given an ambiguity rate *m* in *Lex*, the derivation procedure would have an exponential order of complexity $O(m^n)$. We can mitigate this, either by selecting the element(s) bringing only coherent (i.e. expected) categories (a categorial priming strategy, Ziegler et al., 2019) or to use a statistical oracle, following Stabler (2013), to limit (or rank) the number of possible alternatives. It is however important to stress that lexical ambiguity is the major source of complexity in this derivation: syntactic ambiguity is greatly subsumed by the lexicon, being the source of structural differences related to the set of categorial expectations processed and to the order in which lexical items are introduce in the derivation. With the strict version of MERGE defined in (15), no attachment ambiguity is allowed, since a matching selection must be readily satisfied as soon as the relevant configuration is created (but see Chesi & Brattico 2018). This is not the case if we would admit "free merge" instead of select/licensors-driven merge: in the first case, admitting that MERGE($l_{1(S1)}$, $l_{2(L2)}$) is possible also if $S_1 \neq L_2$, would produce a syntactic ambiguity which is (again exponentially) proportional to the number of items merged in the structure. This is a crucial argument to prefer feature-driven Merge. Notice, moreover, that admitting that re-merge is also possible without proper licensors/selectors, would quickly lead to unbounded unstoppable recursion. This must be prevented if we want to avoid the *halting* problem. Therefore, the licensors/selectors option seem to be a more logical, self-contained, solution.

[8] We have various options to implement this selection: a specific feature (+focus, +topic, +presupposed etc.) can be added to the relevant item (but this would lead to a proliferation of lexical ambiguity, e.g. [$_{D}$ the …] vs [$_{FOC\ D}$ the …]) or we assume that certain superordinate items can select specific categories, without deleting them (e.g. [$_{+D}$ ε $_{FOC}$]). In this implementation, I will pursue this second, more economic, alternative.

# 4 Parameters

A set of parameters can extend the power of e-MGs in a relevant way both excluding unwelcome structures (non-agreeing constituents) or including various kinds of "discontinuous" phrasal structures that cannot be implemented in a (explanatorily) satisfactory way, but that are attested in different languages. *Parameters* minimally operate on *F* and *Cat* to show the impact of various linguistic assumptions, in terms of generative power, without altering the general architecture of G and the dynamics of the derivation. Below two possible parameters dealing with Agreement and "reconstruction". While the first does not increase the power of the grammar, the second do (see "delay phase projection" discussion in Chesi, 2007, also referred as "delayed phase expansion" or simply "late-expansion" as alternative to "sinking", Chesi & Brattico, 2018). A relevant set of cases is presented to exemplify the behavior of the e-MG derivation.

## 4.1 Agree categories

Agreement is a cross-linguistically parametrized option inducing specific featural unification between two distinct items. A list of categories requiring agreement is provided in the *P(arameters)* set of an e-MG, as well as the specific conditions in which agreement holds. For instance, in Italian, as in many other Romance languages, DPs fully agree in gender and number. To express this, we include *gen(der)* and *num(ber),* in association to *D*, *A*[9] and *N* categories in *Agr* (henceforth, *Agr* features in *l,* e.g. [$_{num.sing,\ gen.fem}$], are abbreviated, i.e. [$_{s,\ f}$]):

(21) *P.Agr = {D.{num, gen}, A.{num, gen}, N.{num, gen}}*

This is sufficient to accept (22).a but not (22).b:

| | | | | |
|---|---|---|---|---|
| (22) | a. | La | prima | notizia |
| | | [$_{D;\ s,\ f}$ the] | [$_{A;\ s,\ f}$ first] | [$_{N;\ s,\ f}$ news] |
| | b. | La | *prime | notizia |
| | | [$_{D;\ s,\ f}$ the] | [$_{A;\ p,\ f}$ first] | [$_{N;\ s,\ f}$ news] |

Similarly, subject-verb agreement and object-past participle (*V.pp*) agreement is expressed as follows:

(23) *P.Agr* = {*T.{per*, *num}*, *V.pp.{num,gen}*}

In SVO languages, S will first be licensed higher than T (unless aux-S inversion applies), then T-S agreement should be checked (*case checking*), then the subject S should reach the thematic role. These three operations are implemented simply including the relevant features in the lexicon as in (24):

(24) [$_{C}$ ε $_{+D,\ =T}$], [$_{T;\ 3,\ s}$ ha $_{+D,\ =V}$], [$_{V}$ cantato $_{=D}$],
*has* *sung*
[$_{D;\ 3,\ s}$ Maria]
*Mary*

Exemplifying the derivation (following the procedure presented in §3), the root node C (phonetically empty) is selected first as the current node *cn* (initialization step), then [$_{D}$ Maria] is merged, satisfying the *+D* expectation of C[10]. The *expect* feature

[9] The nature of adjectival modification cannot be fully addressed here. For simplicity, we assume that intersective adjectives (e.g. "beautiful", as well as restrictive relative clauses and adverbial adjuncts) get licensed by the superordinate item without expectation, while others (e.g. "ordinals") are expected by D (we can use more precise categories following the cartographic approach, Cinque, 2002). This induces a tolerable level of lexical ambiguity (either we assume [$_{D}$ the $_{=A}$], [$_{D}$ the $_{=N}$] or [$_{A}$ ε $_{=N}$]; the second option, [$_{D}$ the $_{=A}$] + [$_{A}$ ε $_{=N}$] in case of DP only composed by D and N, seems more coherent with the cartographic intuitions, and reduces lexical ambiguity).

[10] This instantiates the topic of the predication in a general sense. The features on C can be parametrized: with the +D

+*D* does not delete the *expected* D feature (see note 8), therefore [$_{D\,...}$ Maria] is inserted in the memory buffer of C (since *Maria .expected = D*). [$_{T\,...}$ ha $_{+D,\,=V}$] is then merged, satisfying the last expectation of C (i.e., =T). Since T is the last expected item, it inherits the content of the superordinate memory buffer C (i.e. [$_{D}$ Maria]).

In virtue of the +D expectation of T, [$_{D}$ Maria] is remerged and since both T and D categories are in *P.Agr*, agreement must be verified between [$_{T;\,3,\,s}$ ha] and [$_{D;\,3,\,s}$ Maria]. The check is successful, but still [$_{D\,...}$ Maria] remains in memory (because, again, of the +*D* expectation of T), and it is transmitted to V, which, in the end, is the last expected category of T. Now the =*D* expectation of V finally removes [$_{D}$ Maria]) from memory and licenses it as a V first ("external") argument.

On a similar vein we can implement object clitic – past participle agreement. Notice however that the simple specification of the relevant categories in *P.Agr* would predict that the past participle always agrees with the object, also when it just appears in a post-verbal position. This is an incorrect prediction as shown in (25).b:

(25) a. Maria l'ha cantata
M. it$_{CL f.s}$ has sung$_{f.s}$
b.*M. ha cantata una canzone
M. has sung$_{f.s}$ [$_{f.s.}$ a song]
b'. M. ha cantato una canzone
M. has sung$_{m.s}$ [$_{f.s.}$ a song]

To capture this, we need to restrict (certain) agreement configurations to elements that are moved/remerged, namely *V.pp* will be an agreement category only when merged with an item taken from memory (i.e., the clitic in (25).a).

We express this by adding a superscript in *P.Agr* relevant categories: i.e. $V.pp^{M}$. It is important to consider sub-specifications of V since we don't want V to agree with the (external) argument of an unergative predicate, (26).a' vs (26).b:

(26) a. Maria ha corso.
M.$_{f,s}$ has run$_{m,s}$
a'.*Maria ha corsa.
M.$_{f,s}$ has run$_{f,s}$
b. Maria è caduta.
M.$_{f,s}$ is fallen$_{f,s}$
*M. has fallen*

This can be captured, not only by marking those inflections in which the relevant agreement features are overt (i.e. *V.pp*, namely past participle) but also considering that the external argument and the internal one are licensed by two different categories, *v* and *V* respectively (Kratzer, 1996), and only the second is relevant in terms of agreement (this is also necessary for selecting the correct auxiliary, have, (26).a, vs be, (26).b).

### 4.2 Delayed expectation

Both remnant movement (Haegeman, 2000), was-für Split (Brattico & Chesi, 2020) and reconstruction (Bianchi & Chesi, 2014) seem to require some sort of "late expansion" of some complement. When the "delayed expectation" parameter is on, this becomes an option, and an expectation (possibly nested) can be procrastinated. If the item bearing such expectation has only one expect feature, the only available possibility is to wait for its remerge and then expanding such expectation at that time. Certain (non-presuppositional) subjects that do not behave as islands and seem transparent to sub-extraction, require this option to be active. A significant contrast is reported in (27):

(27) [$_{P}$ Of which sculpture] is [$_{D}$ one copy _$_{P}$]
a. *absolutely [perfect _$_{D}$]?
b. already [available _$_{D}$] ?

In (27).a the subject [one copy $_{=P}$] is expected outside the predicative nucleus [perfect] (presuppositional subject) and there it can't receive its argument [$_{P}$ of which masterpiece] (it is in a nested position). In (27).b, reconstruction is possible under the stage level predicate [available], but the P expectation of [one copy $_{=P}$] must wait to be fulfilled after the subject is reconstructed as an argument of the predicate.
Similarly, to capture the relevant dependency in inverse copular constructions we need this option:

(28) La causa della rivolta sono le foto del muro
The cause of the riot are the pictures of_the wall

According to (Moro, 1997) [$_{D}$ cause] is in fact the predicate and [$_{D}$ picture], the subject of such predicate. To integrate *picture* in the correct position we need to include the relevant expectation under the predicate, i.e., [$_{D}$ cause $_{=D,\,=P}$][11], then to wait for the feature associated to C (or below) we obtain the SV parameterization (which is different from the classic *head directionality* parameter).

[11] Being the subject the "external argument" it should come first than =P, which is the expectation triggering Merge of the "internal argument" [$_{P}$ of the riot].

=D projection (*delayed expectation*) after the predicate is remerged after the copula (that selects a D qualifying as predicate, that is, bringing another D expectation). Notice that while agreement parameterization decreases the derivational complexity (restricting the set of successful merges), *delayed expectation* introduces and extra level of syntactic ambiguity that it proportional to the number of expectations of each lexical item (Chesi, 2007).

## 5 Generation and Parsing Tasks

It is worth to remind to the reader that so far, we just discussed a general derivation which was not implemented as a "performance" task. That means that both a Generation and a Parsing procedure must be defined.

### 5.1 Generation

As far as Generation is concerned, the procedure described in §3 is in fact integrally adopted and it is sufficient to produce the expected sentence with the associated, dependency-based, structural description. As long as the sequence of words *w* is concerned, once a root node is selected, it is easy to imagine a dynamic function, instead of the static ordered sequence *w*, that incrementally proposes items to be integrated, given the history of the derivation or, at least, the last expectation (a sort of structural priming, possibly enriched with semantic features if we add to the lexicon *Sem(antic)* specifications in addition to *Cat* and *Phon* ones). Notice that the lexicon can include phonetically empty categories; this is not a problem for the generation procedure, that consumes input tokens one by one, and then considers a phonetically empty category on a par with phonetically realized ones, namely each item should be present as an incoming token to be processed.

### 5.2 Parsing

As long as phonetically empty items are concerned, the Parsing procedure is minimally different since it must postulate these items (e.g. in prodrop languages), by deducting that the *w* sequence received in input is incomplete/incompatible with certain structural hypotheses. One proposal (Brattico & Chesi 2020) relies on inflectional morphology as an overt realization of unambiguous person and number features cliticized on the predicate, hence doubling the (null) subject. Otherwise, only after a relevant category is selected (with its agreement features) and unmatched by the current input, the empty item could be postulated. This non-determinism is exacerbated by the attachment/selection ambiguity: given $[l_{1\ =/+X}\ [l_{2\ =/+X}]]$, for instance, an incoming item with *X exp(ect)ed* feature that should be merged with $l_2$ first, according to the derivation algorithm provided in §3, could, in fact, be merged also with $l_1$, assuming that $l_{2\ =X}$ expectation can be satisfied with an empty item bearing *X* as *exp(ect)ed*. Similarly, an adjunct marked with *Y exp(ect)ed* category could be merged with both $l_1$ and $l_2$ in $[l_1\ [l_2]]$ in case of lexical ambiguity ($[l_1]$, $[l_{1\ +Y}]$, $[l_2]$, $[l_{2\ +Y}]$). In this sense, the derivation procedure in §3 is insufficient as a full-fledged parsing strategy and must be integrated with disambiguation routines dealing with the possibilities just mentioned. It is important to stress that these disambiguation strategies do not alter the general derivation procedure introduced here, which remains the lowest common denominator of Generation and Parsing in e-MGs. The relation between grammar and parser (and, more generally, competence and performance) is monotonic.

## 6 Conclusions

The e-MGs formalization proposed here is a simple (parametrized) framework suitable for comparing syntactic (competence-based) predictions and human parsing/generation performance. This is made possible by the core derivation assumed, which is the same in both tasks (back to the *token transparency* hypothesis discussed in Miller & Chomsky 1963). While there is little to add to implement a full-fledged Generation procedure (see §3.2), as long as the Parsing perspective is concerned, the information asymmetry of this task with respect to Generation requires extra routines to be implemented, in addition to the basic derivation algorithm: lexical ambiguity must be resolved "on-line" and phonetically empty items must be postulated when needed. This creates an extra level of complexity which is however manageable under the same derivational perspective here presented: the core derivation is sufficiently specified to operate independently from parsing-specific disambiguation assumptions which operate monotonically with respect to MERGE, MOVE and AGREE. This is an ideal foothold for metrics that aim at comparing the predicted difficulty not only globally (De Santo, 2020; Graf et al., 2017) but also "on-line" that is, on a word by word basis (Chesi & Canal, 2019) as illustrated in the attached simple implementation.

**Implementation:**
https://github.com/cristianochesi/e-MGs

## References

Alexiadou, A., & Anagnostopoulou, E. (1998). Parametrizing AGR: Word order, V-movement and EPP-checking. *Natural Language & Linguistic Theory*, *16*(3), 491–539.

Bianchi, V., & Chesi, C. (2006). Phases, left-branch islands, and computational nesting. *Proceedings of the 29th Annual Penn Linguistics Colloquium*, *12.1*, 15–28.

Bianchi, V., & Chesi, C. (2014). Subject islands, reconstruction, and the flow of the computation. *Linquistic Inquiry*, 525–569. https://doi.org/10.1162/LING_a_00166

Brattico, P., & Chesi, C. (2020). A top-down, parser-friendly approach to pied-piping and operator movement. *Lingua*, *233*(102760), 1–28. https://doi.org/10.1016/j.lingua.2019.102760

Chesi, C. (2005). Phases and Complexity in Phrase Structure Building. In *Computational Linguistics in the Netherlands 2004: Selected Papers of the 15th Meeting of Computational Linguistics in the Netherlands* (pp. 59–75). LOT. http://lotos.library.uu.nl/publish/issues/4/

Chesi, C. (2007). An introduction to Phase-based Minimalist Grammars: Why move is Top-Down from Left-to-Right. In *STIL - Studies in Linguistics—Vol. 1* (Vol. 1, pp. 38–75). CISCL Press.

Chesi, C. (2015). On directionality of phrase structure building. *Journal of Psycholinguistic Research*, 65–89. https://doi.org/10.1007/s10936-014-9330-6

Chesi, C. (2021). Expectation-based Minimalist Grammars. *ArXiv:2109.13871 [Cs]*. http://arxiv.org/abs/2109.13871

Chesi, C. (2018). An efficient Trie for binding (and movement). *Proceedings of the Fifth Italian Conference on Computational Linguistics (CLiC-It 2018)*, *2253*. https://www.scopus.com/inward/record.uri?eid=2-s2.0-85057729135&partnerID=40&md5=3c941a7524597857a24b64d671e7239a

Chesi, C., & Brattico, P. J. (2018). Larger than expected: Constraints on pied-piping across languages. *RGG. RIVISTA DI GRAMMATICA GENERATIVA*, *2008.4*, 1–38.

Chesi, C., & Canal, P. (2019). Person Features and Lexical Restrictions in Italian Clefts. *FRONTIERS IN PSYCHOLOGY*. https://doi.org/10.3389/fpsyg.2019.02105

Chomsky, N. (1981). *Lectures on government and binding: The Pisa lectures*. Walter de Gruyter.

Chomsky, N. (1995). *The minimalist program*. MIT press.

Chomsky, N. (2001). Derivation by phase. In M. Kenstowicz (Ed.), *Ken Hale: A life in language* (pp. 1–52). MIT Press.

Chomsky, N. (2008). On phases. In R. Freidin, C. P. Otero, & M. L. Zubizarreta (Eds.), *Foundational issues in linguistic theory: Essays in Honor of Jean-Roger Vergnaud* (Vol. 45, pp. 133–166). MIT Press.

Chomsky, N. (2020). *UCLA Lectures*. lingbuzz/005485

Chomsky, N., Gallego, Á. J., & Ott, D. (2019). Generative grammar and the faculty of language: Insights, questions, and challenges. *Catalan Journal of Linguistics*, 229–261.

Cinque, G. (2002). *Functional Structure in DP and IP: The Cartography of Syntactic Structures, Volume 1*. Oxford University Press.

De Santo, A. (2020). MG Parsing as a Model of Gradient Acceptability in Syntactic Islands. *Proceedings of the Society for Computation in Linguistics 2020*, 59–69. https://www.aclweb.org/anthology/2020.scil-1.7

Frazier, L., & Clifton, C. (1989). Successive cyclicity in the grammar and the parser. *Language and Cognitive Processes*, *4*(2), 93–126. https://doi.org/10.1080/01690968908406359

Graf, T., & Marcinek, B. (2014). Evaluating evaluation metrics for minimalist parsing. *Proceedings of the Fifth Workshop on Cognitive Modeling and Computational Linguistics*, 28–36.

Graf, T., Monette, J., & Zhang, C. (2017). Relative clauses as a benchmark for Minimalist parsing. *Journal of Language Modelling*, *5*(1). https://doi.org/10.15398/jlm.v5i1.157

Haegeman, L. (2000). Remnant movement and OV order. In P. Svenonius (Ed.), *The derivation of VO and OV* (pp. 69–96). John Benjamins Publishing Company Amsterdam.

Hale, J. (2001). A Probabilistic Earley Parser as a Psycholinguistic Model. *Second Meeting of the North American Chapter of the Association for Computational Linguistics*. https://aclanthology.org/N01-1021

Hale, J. (2011). What a rational parser would do. *Cognitive Science*, *35*(3), 399–443.

Heim, I., & Kratzer, A. (1998). *Semantics in generative grammar*. Blackwell.

Huang, C.-T. J. (1982). *Logical relations in Chinese and the theory of grammar*. MIT.

Kayne, R. S. (1994). *The antisymmetry of syntax*. MIT Press.

Kayne, R. S. (2020). Antisymmetry and Externalization. *Ms. New York University*. https://doi.org/lingbuzz/005554

Kobele, G. M., Gerth, S., & Hale, J. (2013). Memory Resource Allocation in Top-Down Minimalist Parsing. In G. Morrill & M.-J. Nederhof (Eds.), *Formal Grammar* (Vol. 8036, pp. 32–51). Springer Berlin Heidelberg. https://doi.org/10.1007/978-3-642-39998-5_3

Kratzer, A. (1996). Severing the External Argument from its Verb. In J. Rooryck & L. Zaring (Eds.), *Phrase Structure and the Lexicon* (Vol. 33, pp. 109–137). Springer Netherlands. https://doi.org/10.1007/978-94-015-8617-7_5

Levy, R. (2008). Expectation-based syntactic comprehension. *Cognition*, *106*(3), 1126–1177.

Levy, R., & Keller, F. (2013). Expectation and locality effects in German verb-final structures. *Journal of Memory and Language*, *68*(2), 199–222. https://doi.org/10.1016/j.jml.2012.02.005

Michaelis, J. (2001). Derivational Minimalism Is Mildly Context–Sensitive. In M. Moortgat (Ed.), *Logical Aspects of Computational Linguistics* (Vol. 2014, pp. 179–198). Springer Berlin Heidelberg. https://doi.org/10.1007/3-540-45738-0_11

Miller, G. A., & Chomsky, N. (1963). Finitary Models of Language Users. In D. Luce (Ed.), *Handbook of Mathematical Psychology* (pp. 2–419). John Wiley & Sons.

Momma, S., & Phillips, C. (2018). The Relationship Between Parsing and Generation. *Annual Review of Linguistics*, *4*(1), 233–254. https://doi.org/10.1146/annurev-linguistics-011817-045719

Moro, A. (1997). *The raising of predicates: Predicative noun phrases and the theory of clause structure* (Vol. 80). Cambridge University Press.

Nivre, J., Agić, Ž., Ahrenberg, L., Antonsen, L., Aranzabe, M. J., Asahara, M., Ateyah, L., Attia, M., Atutxa, A., Augustinus, L., & others. (2017). *Universal Dependencies 2.1*.

Nunes, J., & Uriagereka, J. (2000). Cyclicity and Extraction Domains. *Syntax*, *3*(1), 20–43. https://doi.org/10.1111/1467-9612.00023

Partee, B. H., Meulen, A. ter, & Wall, R. E. (1993). *Mathematical methods in linguistics* (Corrected second printing of the first edition). Kluwer Academic Publishers.

Pollard, C. J., & Sag, I. A. (1994). *Head-driven phrase structure grammar*. Center for the Study of Language and Information ; University of Chicago Press.

Rizzi, L. (2016). Labeling, maximality and the head–phrase distinction. *The Linguistic Review*, *33*(1), 103–127.

Ross, J. R. (1967). *Constraints on variables in syntax.* MIT.

Stabler, E. (2011). Computational Perspectives on Minimalism. In C. Boeckx (Ed.), *The Oxford Handbook of Linguistic Minimalism*. Oxford University Press. https://doi.org/10.1093/oxfordhb/9780199549368.013.0027

Stabler, E. (2013). Two Models of Minimalist, Incremental Syntactic Analysis. *Topics in Cognitive Science*, *5*(3), 611–633. https://doi.org/10.1111/tops.12031

Stabler, E. (1997). Derivational minimalism. In C. Retoré (Ed.), *Logical Aspects of Computational Linguistics* (pp. 68–95). Springer Berlin Heidelberg.

Ziegler, J., Bencini, G., Goldberg, A., & Snedeker, J. (2019). How abstract is syntax? Evidence from structural priming. *Cognition*, *193*, 104045. https://doi.org/10.1016/j.cognition.2019.104045